\documentclass[letterpaper, 10 pt, conference]{ieeeconf}  % Comment this line out if you need a4paper

\IEEEoverridecommandlockouts                              % This command is only needed if 
\title{\LARGE \bf
Grounded Vision-Language Interpreter for \\Long-Horizon Bimanual Task and Motion Planning
}

\author{
Jeremy Siburian$^{1,2,*}$ \quad Keisuke Shirai$^{3,*,\dagger}$ \quad Cristian C. Beltran-Hernandez$^{1,*,\ddagger}$\\
  Masashi Hamaya$^{1}$ \quad Michael Görner$^{4}$ \quad Atsushi Hashimoto$^{1}$\\
  $^{1}$OMRON SINIC X Corporation, $^{2}$The University of Tokyo, $^{3}$AIST, $^{4}$Constructor University\\
  \thanks{This work was supported by JSPS KAKENHI Grant Number 25K21274.}
  \thanks{$^*$Equal contribution. $^\dagger$Work done at OMRON SINIC X Corporation.}
  \thanks{$^\ddagger$Corresponding author: cristian.beltran@sinicx.com}
}

\usepackage{amsmath,amsfonts,bm}

\def\eqref#1{equation~\ref{#1}}

\def\1{\bm{1}}

\DeclareMathAlphabet{\mathsfit}{\encodingdefault}{\sfdefault}{m}{sl}
\SetMathAlphabet{\mathsfit}{bold}{\encodingdefault}{\sfdefault}{bx}{n}

\def\gA{{\mathcal{A}}}

\def\gD{{\mathcal{D}}}

\def\gG{{\mathcal{G}}}

\def\gI{{\mathcal{I}}}

\def\gL{{\mathcal{L}}}
\def\gM{{\mathcal{M}}}

\def\gO{{\mathcal{O}}}
\def\gP{{\mathcal{P}}}

\def\gS{{\mathcal{S}}}

\usepackage{amssymb}
\usepackage{algorithm}
\newenvironment{tightlist}%
{\begin{list}{$\bullet$}{%
    \setlength{\topsep}{0in}
    \setlength{\partopsep}{0in}
    \setlength{\itemsep}{0in}
    \setlength{\parsep}{0in}
    \setlength{\leftmargin}{1.5em}
    \setlength{\rightmargin}{0in}
}
}%
{\end{list}
}

\usepackage{booktabs}
\usepackage{graphicx}
\usepackage{comment}
\usepackage{float}
\usepackage{mathtools}
\usepackage{float}
\usepackage[normalem]{ulem}
\usepackage{listings}
\usepackage{color}
\usepackage{multicol}
\usepackage{footnote}
\usepackage[table,xcdraw,dvipsnames]{xcolor}
\usepackage{todonotes}

\usepackage[noend,]{algpseudocode}
\usepackage{fancyvrb}

\usepackage{hyperref}
\usepackage{dblfloatfix}
\usepackage{graphicx} 
\usepackage{multicol}
\usepackage{multirow}
\usepackage{listings}
\usepackage{xcolor}

\newcommand{\pddl}[1]{{\texttt{#1}}} %

\definecolor{LightGreen}{RGB}{226, 246, 211}
\definecolor{DarkGreen}{RGB}{184, 233, 148}
\definecolor{MyDarkBlue}{rgb}{0,0.08,0.8}
\definecolor{MyDarkGreen}{RGB}{45,155,45}
\definecolor{MyDarkRed}{rgb}{0.8,0.02,0.02}
\definecolor{MyDarkOrange}{rgb}{0.40,0.2,0.02}
\definecolor{MyPurple}{RGB}{111,0,255}
\definecolor{MyRed}{rgb}{0.8,0.0,0.0}
\definecolor{MyGold}{rgb}{0.75,0.6,0.12}
\definecolor{MyDarkgray}{rgb}{0.66, 0.66, 0.66}


\usepackage{algorithm}
\usepackage[noend]{algpseudocode}
\usepackage[export]{adjustbox}
\algnewcommand\algorithmicdeclare{\textbf{Assume:}}
\algnewcommand\Declare{\item[\algorithmicdeclare]}

\usepackage{listings}
\let\labelindent\relax
\usepackage{enumitem}
\usepackage{pifont} %
\usepackage{algorithm}
\usepackage[noend]{algpseudocode}

\usepackage{framed}
\definecolor{shadecolor}{gray}{0.95}
\usepackage{xparse}
\usepackage{etoolbox}
\usepackage{setspace}

\usepackage[most]{tcolorbox}
\usepackage{lipsum}
\tcbuselibrary{skins,breakable}
\usetikzlibrary{shadings,shadows}
    {\endtcolorbox}

\definecolor{ColorAskVLM}{RGB}{203, 242, 220}
\definecolor{ColorVLMAnswer}{RGB}{204, 229, 245}
\definecolor{ColorComment}{RGB}{116, 125, 140}
\definecolor{ColorKeyword}{RGB}{87, 96, 111}
\usepackage{listings}
\usepackage{changepage}
\lstdefinestyle{askvlm}{
  rulecolor=\color{ColorAskVLM},
  backgroundcolor=\color{ColorAskVLM},
}
\lstdefinestyle{vlmanswer}{
  rulecolor=\color{ColorVLMAnswer},
  backgroundcolor=\color{ColorVLMAnswer},
}

\usepackage{setspace}
\usepackage{inconsolata}
\usepackage[T1]{fontenc}
\usepackage{xspace}
\usepackage[capitalise,nameinlink]{cleveref}
\crefname{section}{Sect.}{Sect.}
\Crefname{section}{Section}{Sections}
\crefname{figure}{Fig.}{Fig.}
\Crefname{figure}{Figure}{Figures}
\crefname{table}{Tab.}{Tab.}
\Crefname{table}{Table}{Tables}

\usepackage{wrapfig}
\usepackage{tabularx}
\usepackage{arydshln}
\usepackage{etoc}

\definecolor{codebg}{rgb}{0.95, 0.95, 0.92}
\definecolor{codecomment}{rgb}{0.3, 0.6, 0.3}
\definecolor{codekeyword}{rgb}{0.0, 0.3, 0.7}
\definecolor{codestring}{rgb}{0.58, 0, 0.82}

\lstnewenvironment{speciallstlisting}{
    \lstset{
        language=Python, % Specify the programming language here
        backgroundcolor=\color{codebg},
        basicstyle=\ttfamily\small,
        keywordstyle=\color{codekeyword},
        commentstyle=\color{codecomment},
        stringstyle=\color{codestring},
        numbers=left,
        numberstyle=\tiny\color{codecomment},
        stepnumber=1,
        frame=single,
        framesep=5pt,
        framerule=0pt,
        rulecolor=\color{codecomment},
        tabsize=4,
        showstringspaces=false,
        breaklines=true,
        breakatwhitespace=true,
        captionpos=b
    }
}{}

\newcommand{\ignore}[1]{}

\makeatletter
\DeclareRobustCommand\onedot{\futurelet\@let@token\@onedot}
\def\@onedot{\ifx\@let@token.\else.\null\fi\xspace}

\makeatother

\definecolor{MyDarkBlue}{rgb}{0,0.08,1}
\definecolor{MyDarkGreen}{rgb}{0.02,0.6,0.02}
\definecolor{MyDarkRed}{rgb}{0.8,0.02,0.02}
\definecolor{MyDarkOrange}{rgb}{0.40,0.2,0.02}
\definecolor{MyPurple}{RGB}{111,0,255}
\definecolor{MyRed}{rgb}{1.0,0.0,0.0}
\definecolor{MyGold}{rgb}{0.75,0.6,0.12}
\definecolor{MyDarkgray}{rgb}{0.66, 0.66, 0.66}
\definecolor{YangRed}{RGB}{231, 76, 60}
\definecolor{YangGreen}{RGB}{39, 174, 96} 
\definecolor{YangBlue}{RGB}{52, 152, 219}
\definecolor{YangOrange}{RGB}{230, 126, 3}
\definecolor{YangPurple}{RGB}{142, 68, 173}

\newif\ifpropositionfirstitem
\propositionfirstitemtrue

\definecolor{light-gray}{rgb}{0.8, 0.8, 0.8}
\definecolor{highlight}{HTML}{e3eeff}
\newtcolorbox{HLPromptBox}{%
  enhanced,
  breakable,
  width=0.96\linewidth,          % always as wide as current line
  colback=highlight,
  colframe=highlight,
  boxrule=0pt,
  sharp corners,
  left=2pt, right=2pt, top=1pt, bottom=1pt,
  before skip=0pt, after skip=0pt,
  fontupper=\strut            % identical height/depth each line
}

\newcommand{\vilain}{ViLaIn\xspace}
\newcommand{\vilaintamp}{ViLaIn-TAMP\xspace}

\lstdefinelanguage{pddl}{
    basicstyle=\ttfamily\fontsize{8}{10}\selectfont,
    sensitive=true,
    morecomment=[l]{;}, 
    morestring=[b]",
    keywords={define,domain,problem,agent,object,requirements,types,predicates,action,parameters,precondition,postcondition,goal,init,forall,exists,and,or,not,imply,when,either,effect},
    keywordstyle=\color{blue!70!black}\bfseries,
    morekeywords={[2]},
    keywordstyle={[2]\color{green!40!black}\bfseries},
    morekeywords={[3]},
    keywordstyle={[3]\color{orange!80!black}\bfseries},
    basicstyle=\ttfamily\fontsize{10}{12}\selectfont,
    commentstyle=\color{gray}\itshape,
    stringstyle=\color{red!70!black},
    showstringspaces=false,
    numbers=left,
    numberstyle=\tiny\color{gray},
    numbersep=8pt,
    tabsize=2,
    breaklines=true,
    breakatwhitespace=false,
    backgroundcolor=\color{gray!10},
    captionpos=b,
    belowcaptionskip=0.5em,
    escapeinside={(*@}{@*)},
}

\newcommand{\figuretextspace}{\vspace{-2.8mm}}

\begin{document}

\maketitle
\thispagestyle{empty}
\pagestyle{empty}

%%%%%%%%%%%%%%%%%%%%%%%%%%%%%%%%%%%%%%%%%%%%%%%%%%%%%%%%%%%%%%%%%%%%%%%%%%%%%%%%
\begin{abstract}
While recent advances in vision-language models have accelerated language-guided robot planning, their black-box nature lacks the safety guarantees and interpretability crucial for real-world deployment. Conversely, classical symbolic planners offer rigorous safety verification but require significant expert knowledge for setup. Moreover, most existing methods are limited to single-arm pick-and-place tasks, leaving bimanual manipulation largely underexplored, despite its tightly interdependent subtasks and the need to explicitly manage inter-arm collisions. To bridge this gap, this paper proposes \vilaintamp, a hybrid planning framework for enabling verifiable, interpretable, and autonomous bimanual robot behaviors. \vilaintamp comprises three main components: (1) a Vision-Language Interpreter (\vilain) adapted from a prior work that converts multimodal inputs into structured PDDL problem specifications, (2) an integrated Task and Motion Planning (TAMP) system that grounds these specifications in actionable trajectory sequences through symbolic and geometric constraint reasoning, explicitly verifying feasibility before execution, and (3) a corrective planning (CP) module which receives structured motion failure feedback and feeds it as constraints back to \vilain to refine the specification. We design challenging bimanual manipulation tasks in a cooking domain to evaluate our framework, where experimental results show that \vilaintamp outperforms a VLM-as-a-planner baseline by 17.5\% in mean success rate, with the CP module boosting it further by 32.9\%. We further validate \vilaintamp on a physical dual-arm robotic system. Project page: \\\url{https://omron-sinicx.github.io/ViLaIn-TAMP}
\end{abstract}
\section{Introduction}
Robots in human-centered environments are expected to plan and execute diverse, complicated tasks autonomously and safely. Instructing such robots through natural language is the long-standing dream of artificial intelligence and robotics research, as natural language not only supports robot execution by non-experts but also enables people to control robots intuitively. These robots are required to interpret linguistic instructions and make feasible plans by perceiving the environment. Large language models (LLMs) and vision-language models (VLMs)~\cite{openai2024gpt4o,meta2024llama3,geminiteam2024gemini15} have the potential to alleviate this barrier. 
\begin{figure}[t]
    \centering
    \includegraphics[width=0.925\linewidth]{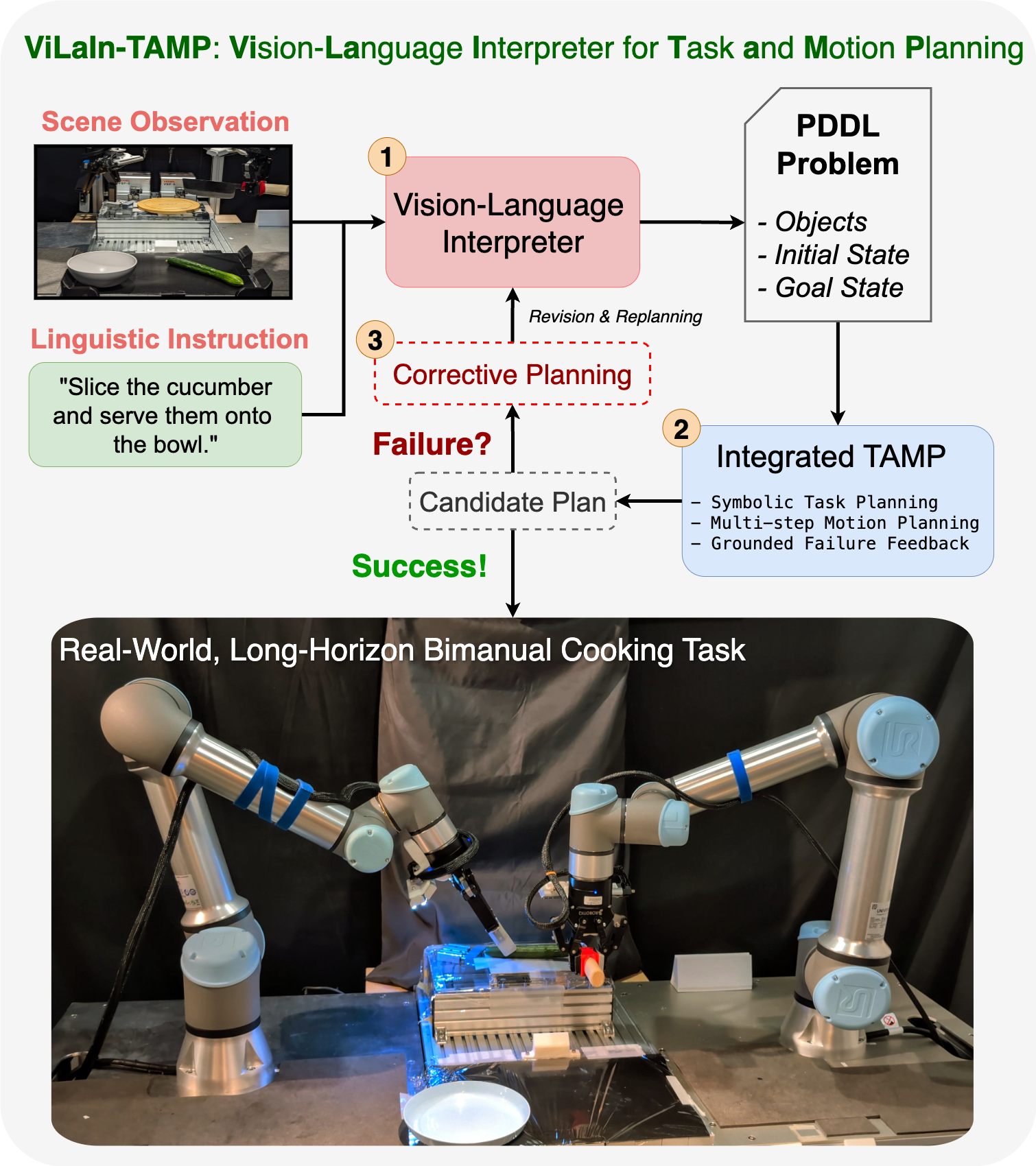}
    \caption{\textbf{\vilaintamp} is a novel end-to-end planning framework for long-horizon bimanual manipulation that 1) converts multimodal inputs into PDDL problems, 2) finds feasible motion plans via an integrated TAMP system, and 3) reasons over failure feedback for corrective replanning.
    }
    \label{fig:main_overview}
    \figuretextspace
\end{figure}
In recent years, research on LLM-based planning has gained attention~\cite{huang2023inner,singh2023progprompt,silver2024generalized}. The LLM/VLM-as-a-planner approach, which directly converts linguistic instructions (and visual observation) into plans, is a representative example, and methods have been proposed in various domains and task settings~\cite{mei2024repanvlm,wang2024llm3,yang2025guiding}. However, their black-box nature has fundamental flaws, notably the missing logical validation and unclear physical feasibility of generated plans. 

Vision-Language Interpreter (\vilain)~\cite{shirai2024vision} addresses the former limitation by performing hybrid planning through a symbolic planner. \vilain first generates PDDL~\cite{fox2003pddl2} problem specifications from the instructions and then triggers a symbolic planner to find plans, while syntactically and semantically validating the specifications. Nonetheless, \vilain is limited to task planning, and it cannot be directly connected to real robot execution. We bridge this gap while addressing the latter limitation by building a novel framework based on \vilain and an integrated TAMP system.

In particular, most existing VLM-based planning methods target single-arm settings, leaving bimanual manipulation and its unique challenges largely underexplored~\cite{zhu2023vima,yang2025guiding}. Consider the dual-arm cooking task shown in~\cref{fig:main_overview} of slicing a vegetable with a knife. This long-horizon task contains subtasks that are strongly interrelated and cannot be considered independently from each other. For instance, one arm must fixture the object while the other slices, and certain grasp poses may cause inter-arm collisions. Furthermore, collisions among grasped objects/tools, workspaces, and robots must be considered for safety, yet most motion planners such as MoveIt~\cite{coleman2014reducing} are limited to pick-and-place pipelines, have black-box implementations, and ignore interdependence between subtasks~\cite{fresnillo2023extending}. We therefore integrate the MoveIt Task Constructor (MTC)~\cite{gorner2019moveit} for multi-stage manipulation, and significantly modify it to provide multi-level structured motion failure feedback for VLM reasoning, enabling effective corrective planning for dual-arm tasks.

We propose \vilaintamp, a novel end-to-end planning framework that \textit{grounds} \vilain in TAMP for bimanual manipulation. As shown in \cref{fig:main_overview}, \vilaintamp first converts linguistic instructions and scene observations into PDDL problems, finds plans by driving a symbolic planner~\cite{helmert2006fast}, explicitly verifies geometric feasibility via the MTC~\cite{gorner2019moveit}, and finally executes the obtained plans on the real robot. To improve robustness, ViLaIn-TAMP contains a \textit{corrective planning} (CP) module that incorporates failure feedback from the TAMP module to address situations where planning fails, such as when PDDL problems are syntactically incorrect or when logical or geometrical constraints are unsatisfiable. We evaluate \vilaintamp on five bimanual cooking tasks, demonstrating strong planning capability through quantitative simulation experiments and real-world validation on a physical dual-arm system.

We construct an open-source evaluation dataset covering a custom PDDL domain, problems, linguistic instructions, and scene observations. Evaluated on this dataset, \vilaintamp shows strong performance over a VLM-as-a-planner baseline with greater interpretability, and CP improves execution robustness. Upon acceptance, we plan to release our dataset, codebase, and custom MTC implementation to support future research and contribute to the robotics and open-source MoveIt community.

\section{Related Work} 
\label{sec:related-work}

\textbf{Foundation Models in Robot Planning:}
Research and development of large language models (LLMs)~\cite{openai2024gpt4,meta2024llama3} and vision-language models (VLMs)~\cite{openai2024gpt4o,geminiteam2024gemini15} have been extensively developed in natural language processing and computer vision. 
In robotics, the generalization capabilities and web-scale knowledge of these models have led to the development of innovative methods~\cite{kawaharazuka2024real}.
Especially in robot task planning, researchers have developed methods that directly generate action sequence plans~\cite{raman2022planning, mei2024repanvlm}. Prior work showed LLM's strong capability in symbolic planning~\cite{singh2023progprompt,silver2024generalized}, developed mechanisms for replanning from errors using LLMs~\cite{raman2022planning,lin2023text2motion,wang2024llm3}, and realized multimodal planning using VLMs~\cite{mei2024repanvlm,yang2025guiding}. Despite extensive research efforts, the nature of directly generating plans raises issues such as infeasible plans and the lack of interpretability. To address these issues, recent work proposed a hybrid approach of combining LLM/VLMs and symbolic planners~\cite{liu2023llm,xie2023translating,shirai2024vision}, where a human interpretable specification is first generated before planning. 
LLM+P~\cite{liu2023llm} showed the LLM's capability of translating linguistic descriptions into PDDL problem specifications, while Xie et al. (2023)~\cite{xie2023translating} generated PDDL goals from linguistic instructions. Vision-Language Interpreter (\vilain)~\cite{shirai2024vision} extended these ideas by incorporating \textit{multimodality}, enabling whole PDDL problem generation from linguistic instructions and scene observations. However, ViLaIn is currently limited to task planning only and does not consider any real-world robot execution.
Like LLM+P, \vilain follows a \textit{formalizer} paradigm~\cite{liu2023llm, huang2025limit}, translating inputs into a PDDL problem under a given domain. We ground this formalizer in an integrated Task and Motion Planning (TAMP) system and close a feedback loop converting motion-level infeasibility into symbolic constraints for re-formalization.

\begin{figure*}[!t]
    \centering
    \includegraphics[width=0.9\linewidth]{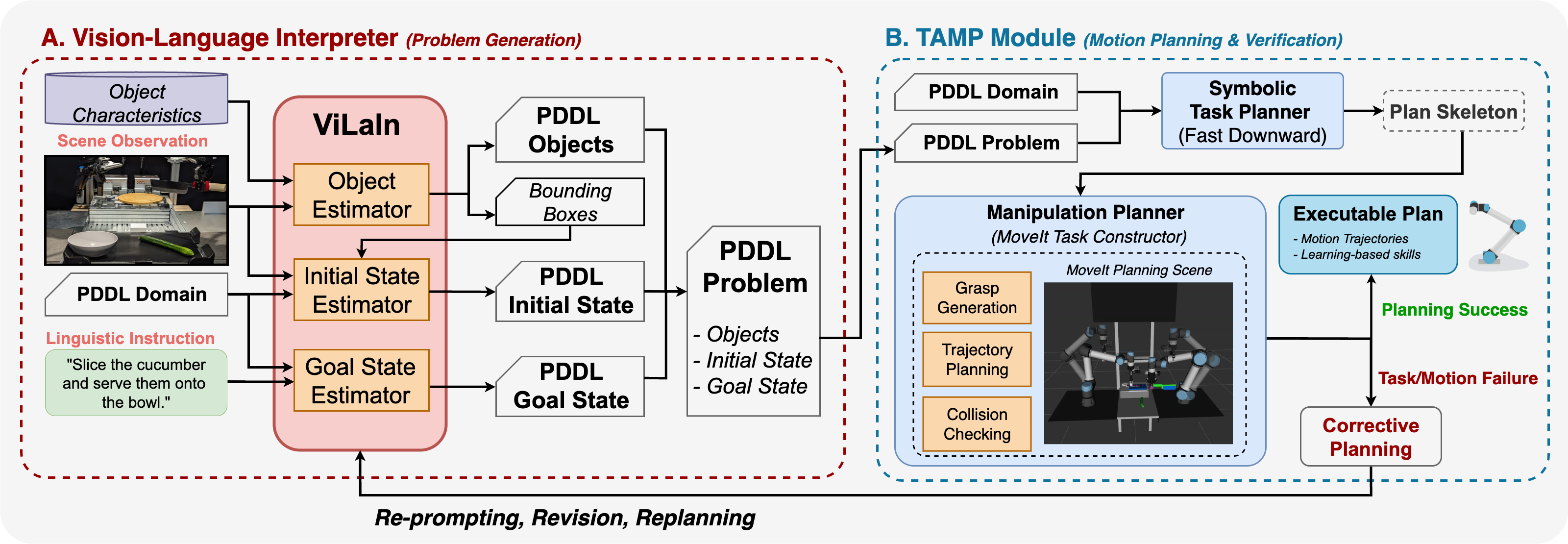}
    \caption{
\textbf{Overview of \vilaintamp framework.} (A) Given a linguistic instruction and scene observation, \vilain generates a complete PDDL problem. (B) The TAMP module solves for symbolic actions and collision-free trajectories. If successful, the plan is executed; otherwise, failures are fed back to \vilain for revision and replanning.
    }
    \label{fig:new_overview}
    \figuretextspace
\end{figure*}

\textbf{LLMs/VLMs for Task and Motion Planning:} 
TAMP planning~\cite{garrett2020pddlstream, siburian2025practical, curtis2022long} typically involves searching for high-level sequences of symbolic actions and sampling for the low-level motion plans that must be executed to achieve a given goal. Recent research has shown promising capabilities in combining large pre-trained models with TAMP~\cite{ding2023task, wang2024llm3, rana2023sayplan}. To improve performance, several works utilized LLMs/VLMs to enforce constraints in TAMP planning and replanning from failure feedback~\cite{chen2024autotamp, guo2025castl}. However, \cite{chen2024autotamp} is limited to 2D navigation tasks, while \cite{guo2025castl} does not consider geometric constraints from motion planning feedback.
Classical constraint-based TAMP solvers such as PDDLStream~\cite{garrett2020pddlstream} and IDTMP~\cite{dantam2018idtmp} propagate geometric infeasibilities as constraints without a foundation model, but assume a correct, fully specified symbolic problem. Here that problem is VLM-produced and may contain misdetected objects, missing predicates, or hallucinated goals; corrective planning thus addresses errors in the problem's \textit{formalization}, not only its motion-level realization.
End-to-end approaches such as RT-X~\cite{oneill2024openx} and VIMA~\cite{zhu2023vima} demonstrate strong performance through large-scale data collection, but lack explicit symbolic planning, making failure introspection and pre-execution verification fundamentally difficult.
\cite{wang2024llm3, raman2024cape, duan2025aha} implemented motion failure reasoning using LLMs/VLMs to correct robot behaviors, but these are limited to single-arm tasks.
\vilaintamp similarly leverages motion failure feedback, but we instead exploit the failure introspection capabilities of the MoveIt Task Constructor framework~\cite{gorner2019moveit} for dual-arm manipulation,
explicitly verifying geometric feasibility before execution, a capability absent in all aforementioned works.
While the feedback loop is not arm-specific, its \textit{actionable content} is: reports identify the responsible arm and inter-arm collision pairs, letting the symbolic layer reason about arm-role assignment (which arm fixtures vs.\ slices), absent in single-arm planning.
VLM-TAMP~\cite{yang2025guiding} is closest to our work, using VLM-generated subgoals refined by a TAMP planner. However, VLM-TAMP is evaluated only in simulation with no real-robot implementation, and provides only collided object names as failure feedback, whereas we extract multi-level structured feedback from our custom MTC implementation for more effective VLM reasoning. Furthermore, while VLM-TAMP and RT-X~\cite{oneill2024openx} support dual-arm configurations, the arms operate relatively independently without explicit inter-arm collision reasoning. In contrast, \vilaintamp handles tightly coupled bimanual coordination (e.g., one arm fixturing while the other slices) with learning-based skills and is validated on a physical dual-arm system.

\section{Preliminaries}

\textbf{Task and Motion Planning (TAMP):} We adopt the TAMP setting~\cite{garrett2021integrated}, which jointly reasons over a discrete \textit{task} level (which symbolic actions to take, and in what order) and a continuous \textit{motion} level (whether collision-free trajectories realize them). Following common assumptions~\cite{garrett2021integrated, silver2021learning}, we assume known object and robot models with approximate object poses from perception, per-action motion samplers (here, MTC), and a collision checker for pre-execution verification. Unlike in pure task planning, a symbolically valid plan must also admit a feasible motion-level realization.

\textbf{PDDL Formulation:} 
We use the Planning Domain Definition Language (PDDL) \cite{fox2003pddl2}, a standardized and human-readable planning language, to formalize target tasks into a symbolic representation, consisting of a \textit{domain} and \textit{problem}. 
A PDDL \textit{domain} is formalized by a set of predicates $\gP$ and actions $\gA$. A PDDL \textit{problem} can be represented as a tuple $(\gO, \gI, \gG)$ which is defined by a set of initial objects $\gO$, an initial state $\gI$, and a goal state $\gG$. We use predicates to describe \textit{states}, which can be represented as \textit{literals}. For example, we use \pddl{(At ?obj ?loc)} to specify that an object \texttt{?obj} is currently at location \texttt{?loc}. We also model geometric constraints using predicates such as \pddl{(CanNotReach ?robot ?obj ?loc)}, to indicate that a robot cannot reach an object at a particular location. In our Cooking domain, we design several types of \textit{actions} that can induce changes to states, such as \texttt{pick}, \texttt{place}, \texttt{equip-tool}, \texttt{fixture}, \texttt{slice}, \texttt{clean-up}, and \texttt{serve-food}.

\textbf{Multimodal Planning Problem Specification:}
We use \vilain to convert multimodal inputs into PDDL problems. This task is formulated as a multimodal planning problem specification, following previous work~\cite{shirai2024vision}. The inputs are represented as $(\mathcal{L}, \mathcal{S}, \mathcal{D})$. $\mathcal{L}$ is a linguistic instruction, $S$ is a scene observation, and $\mathcal{D}$ is domain knowledge, which includes a PDDL domain $(\gP, \gA)$ and other domain-specific features (e.g., object characteristics). The output is a PDDL problem $(\gO, \gI, \gG)$. The goal of this task is to define $\gM: (\gL, \gS, \gD) \rightarrow (\gO, \gI, \gG)$, and we employ \vilain as $\gM$.

\section{Methodology}
\label{sec:method}

The ViLaIn-TAMP framework has three major components: 1) a vision-language interpreter for PDDL problem generation, 2) a sequence-before-satisfy TAMP module for finding symbolically complete, collision-free action plans based on the generated problems, and 3) a corrective planning module for refining outputs based on grounded failure feedback. \cref{fig:new_overview} presents an overview of our framework. 

\subsection{Vision-Language Interpreter for Problem Generation}
\label{subsec:vilain}

Vision-Language Interpreter (\vilain) converts a linguistic instruction, scene observation, and domain knowledge $(\mathcal{L}, \mathcal{S}, \mathcal{D})$ into a PDDL problem $(\gO, \gI, \gG)$. As shown in part A of \cref{fig:new_overview}, \vilain consists of three modules, each generating $\gO$, $\gI$, and $\gG$ of the PDDL problem, respectively. 
While the original \vilain and other prior works used ICL input-output examples \cite{shirai2024vision, liu2023llm, guo2025castl}, we instead provide short natural language descriptions of each predicate and action in the domain \cite{smirnov2024generating, oswald2024large}. These descriptions play a significant role in replacing ICL examples, and we investigate the effects of using ICL examples in our evaluations.

\textbf{Object Estimator} takes $\mathcal{S}$ and the object characteristics of $\mathcal{D}$ as input and estimates the relevant objects $\gO$. The object characteristics are a list of objects of interest with descriptions (e.g., "plate is a white round plate"). The estimator uses a VLM to generate bounding boxes with object labels directly. $\gO$ is automatically created based on the detected object labels. The bounding boxes are used with the labels in the subsequent initial state estimation. 

\textbf{Initial State Estimator} takes $\mathcal{P}$, $\mathcal{S}$, and bounding boxes with object labels as input and estimates $\mathcal{I}$. The estimator feeds the inputs to a VLM and generates $\mathcal{I}$ directly.

\textbf{Goal State Estimator} takes $\mathcal{P}$ and $\mathcal{L}$ and generates $\gG$.

\begin{figure}
    \centering
    \includegraphics[width=0.9\linewidth]{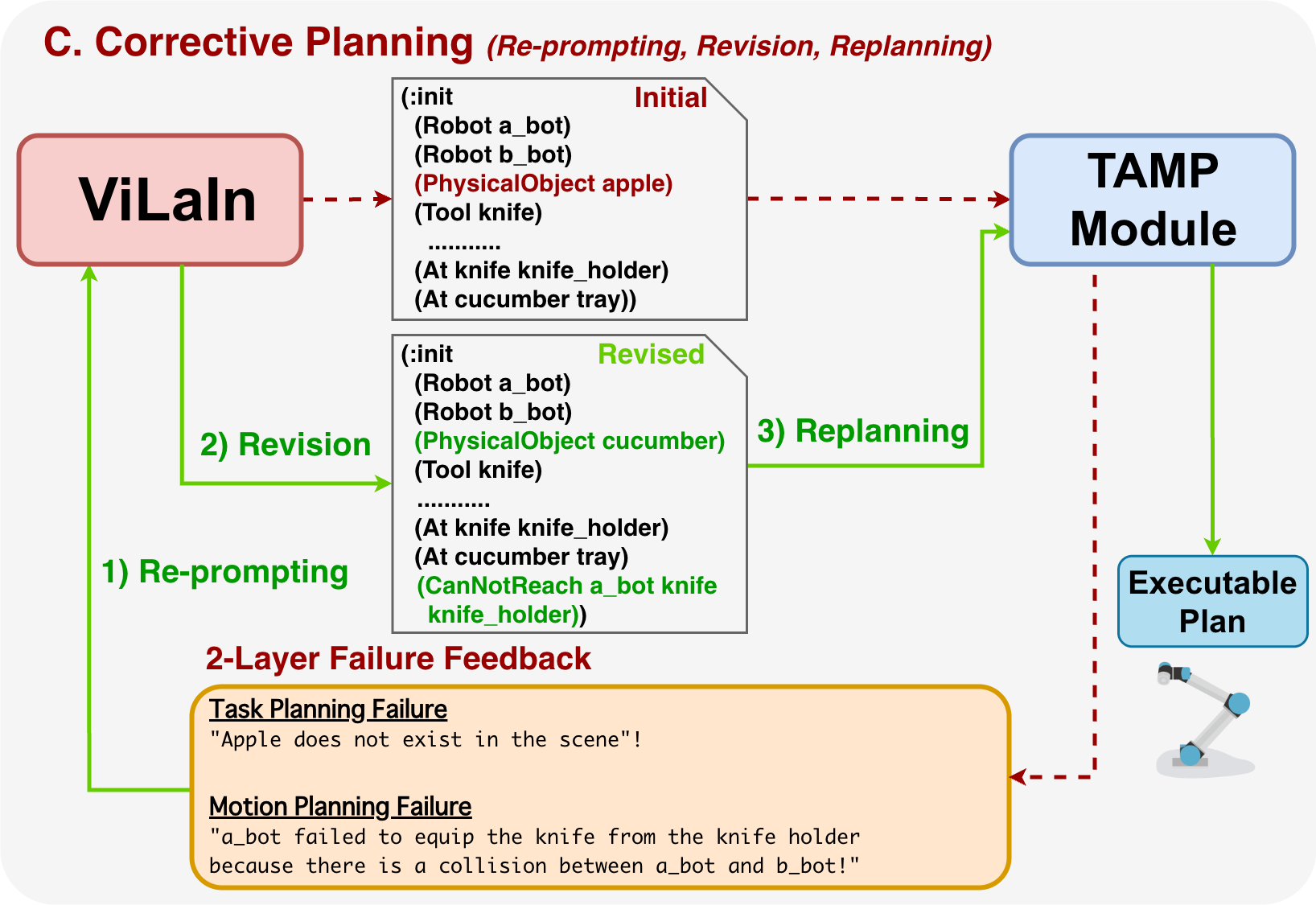}
    \caption{\textbf{Overview of Corrective Planning Module.} \vilaintamp implements a 3-step corrective planning (CP) approach, which involves 1) \textit{re-prompting} the model with the failure feedback to 2) \textit{revise} the PDDL problem, and then 3) \textit{replanning} using the revised PDDL problem.}
    \label{fig:cp_overview}
    \figuretextspace 

\end{figure}

\subsection{Integrated TAMP System}
\label{subsec:tamp}
After a complete PDDL problem is generated by \vilain, an integrated TAMP module is used to search for a symbolically complete, collision-free action plan.

\textbf{Symbolic Task Planning:} For task-level planning, we leverage the off-the-shelf symbolic planner Fast Downward~\cite{helmert2006fast}, which supports PDDL specifications.

\textbf{Multi-step Manipulation Planning:} In complex manipulation tasks, certain stages of a task are often strongly interrelated and cannot be considered independently of each other, particularly for dual-arm manipulation. For example, in the context of our slicing task, certain grasp poses for fixturing an object by an arm may interfere with slicing and cause collisions with the other arm. Conventional motion planners such as MoveIt \cite{coleman2014reducing} are typically restricted to a single-arm pick-and-place pipeline, with limited functions for dual-arm coordination \cite{fresnillo2023extending}. To alleviate this issue, we integrate the MoveIt Task Constructor (MTC) \cite{gorner2019moveit}, an open-source manipulation planning framework for multi-step tasks, into \vilaintamp.
MTC organizes tasks as a hierarchy of stages within serial containers, where each stage passes intermediate states to adjacent stages, naturally capturing interdependencies between manipulation phases.
The framework resolves these interdependencies through controlled co-parameter sampling, supporting finite appropriate attempts for adequate coverage of each symbolic plan.

\begin{figure}[!tp] \centering
  \begin{adjustwidth}{0pt}{1em}
    \centering
    \includegraphics[width=0.8\linewidth]{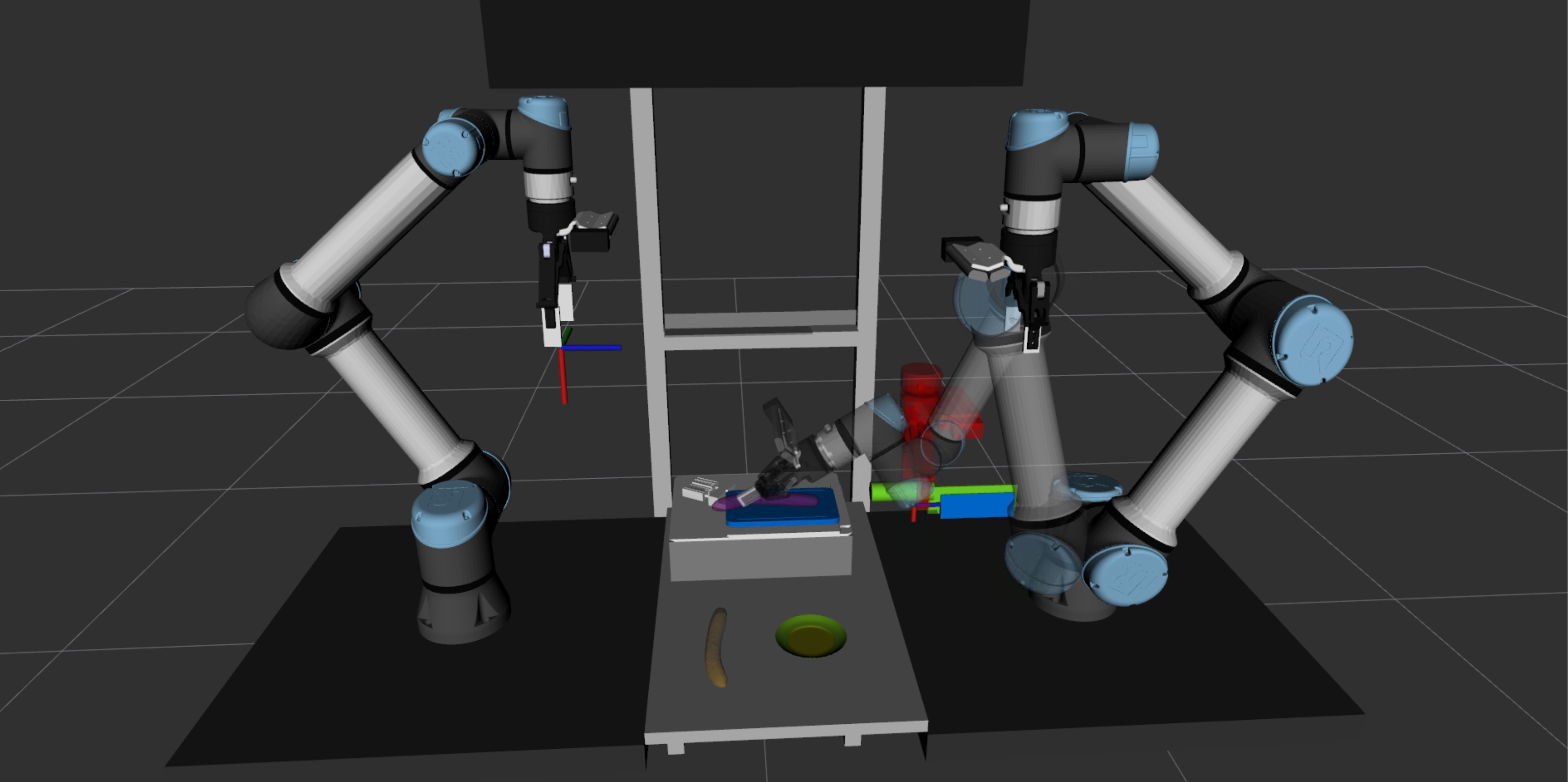}
  \end{adjustwidth}
  \vspace{-5pt}\hspace{0pt}
\begin{adjustwidth}{6pt}{1.7em}
\begin{lstlisting}[style=askvlm]
--- MTC Failure Comments ---
Summary of stages with complete failures:
equip_tool_3 (Pick)
  -> grasp_equip_tool_3 (SimpleGrasp)
    -> compute ik (ComputeIK)
      Failure comments: Collision between 'a_bot_upper_arm_link' and 'b_bot_wrist_2_link'
\end{lstlisting}
\end{adjustwidth}
\vspace{-7pt}\hspace{0pt}
\begin{adjustwidth}{6pt}{1.7em}
\begin{lstlisting}[style=askvlm]
--- Task Execution Trace ---
 1) pick b_bot cucumber tray
 2) place b_bot cucumber cutting_board
 3) fixture b_bot cucumber cutting_board
 4) equip_tool a_bot knife knife_holder cucumber [FAILURE]
 5) slice a_bot knife cucumber cutting_board
\end{lstlisting}
\end{adjustwidth}
\vspace{-7pt}\hspace{0pt}
\begin{adjustwidth}{6pt}{1.7em}
\begin{lstlisting}[style=askvlm]
 --- Personalized Failure Message ---
Motion planning has failed.
Motion failed during the 'equip_tool' action.
During motion planning, a_bot failed to equip the knife from the knife_holder.
From the motion planner feedback, the reason for the action failure is because there is a collision between a_bot_cam_cables_link and b_bot_forearm_link
\end{lstlisting}
\end{adjustwidth}
\vspace{-7pt}\hspace{0pt}
\caption{
\textbf{Motion Planning Failure Feedback.} Our custom MTC extracts multi-level structured feedback (failure comments, execution trace, and synthesized message) for VLM reasoning. Failed plans can also be visualized in RViz before execution.
}
 \label{fig:failure_example}
\figuretextspace
\end{figure}

\textbf{Integrated TAMP Module:} 
We integrated symbolic task planning and manipulation planning using a \textit{sequence-before-satisfy} strategy \cite{garrett2021integrated}. 
As shown in part B of~\cref{fig:new_overview}, we first search for a high-level \textit{sequence} of actions using Fast Downward. After a plan skeleton is found, we use MTC to sample low-level motions that \textit{satisfy} the plan. Each symbolic action is automatically mapped to a corresponding MTC serial container composed of the appropriate stages (e.g., grasp generation, IK computation, trajectory planning, and collision checking), eliminating the need for per-task manual stage specification. Additionally, learning-based skills are integrated into the TAMP framework to handle complex, contact-rich subtasks such as food slicing, which demand rapid adaptation to environmental changes~\cite{beltran2024sliceit}. Following skill-augmented TAMP formulations~\cite{liu2024optimistic, hedegaard2024beyond}, each learned skill is a black-box operator: it exposes PDDL-style pre/post conditions so the task planner sequences it like a primitive action, while geometrically it is defined only by object-centric start/end poses, keeping its policy-driven trajectory opaque. Skills are realized as mock MTC stages; during planning, only trajectories satisfying these pose constraints are valid.

\subsection{Corrective Planning from Failure Feedback}
\label{subsec:cp}
The original \vilain uses a symbolic planner to validate and revise syntactic errors in the generated PDDL problems. However, this ignores geometrical constraints not captured by the symbolic planner, resulting in motion planning failures due to inter-arm collisions, unreachable objects, and nonexistent objects or locations. These errors can be resolved by obtaining feedback from the motion planner and prompting \vilain to revise the PDDLs by adding geometrical constraints (e.g., \texttt{(CanNotReach a\_bot, knife, knife\_holder)}) and modifying incorrect predicates. To realize this feedback loop, we implement a 3-step corrective planning (CP) approach as shown in~\cref{fig:cp_overview}, which involves \textit{re-prompting} the model with the failure feedback to \textit{revise} the PDDL problem, and then \textit{replanning} using the revised PDDL problem. We denote the failure feedback as $\mathcal{E}$. In \vilaintamp, we consider 2-layer feedback from either the task planner or the MTC module.

\textbf{Extracting Motion Failure Feedback for VLM Reasoning:}
MTC provides visual introspection of stage-level failures through its RViz panel, but this information is not programmatically accessible. We modify MTC's C++ API to extract failure information after each planning attempt by traversing the task hierarchy and collecting failure comments from nested stages (e.g., collision pairs from IK computation). We then cross-reference these with the symbolic action sequence to identify the responsible action and robot arm. When motion planning fails, the modified MTC module returns feedback at three levels of abstraction: 1) short failure comments from MTC's internal stage data, 2) the task execution trace indicating at which step failure occurred, and 3) a synthesized natural language message constructed from a template populated with the failed action, robot arm, and specific collision or reachability information. \cref{fig:failure_example} provides an example failure case where a collision occurs between the two robot arms during a food slicing task.

\section{Experimental Evaluation}
\label{sec:experiments}

Our evaluation addresses five questions: ($\mathcal{Q}$1) how effective \vilaintamp is versus a VLM-as-a-planner baseline that directly generates action sequences; ($\mathcal{Q}$2) which failure modes it exhibits; ($\mathcal{Q}$3) how much the CP module improves robustness; ($\mathcal{Q}$4) whether ICL examples are necessary; and ($\mathcal{Q}$5) how the foundation-model choice affects performance.

\subsection{Experiment Setting}
\label{subsec:experiment_setting}

\subsubsection{\textbf{Tasks}} 
We validate our proposed method on five manipulation tasks in the cooking domain:

\begin{tightlist}
    \item \textbf{Pick and Place} aims to move a target object to a desired location.
    \item \textbf{Pick Obstacle Dual Arm} is the same task as Pick and Place except that the location is occupied by another object, causing a collision when directly using the same solution as Pick and Place. One robot removes the collision object (acting as temporary storage) and another arm performs pick-and-place with the target object.
    \item \textbf{Pick Obstacle Single Arm} is the same as Pick Obstacle Dual Arm, but only a single arm is allowed to be used. This task requires more complex planning than Pick Obstacle Dual Arm, as it requires an extra placing action and accurately identifying free placing locations.
    \item \textbf{Slice Food} aims to slice a food item (e.g., fruit or vegetable) using a tool (e.g., knife). A key aspect of the task is that specific choices in which robot to use for certain actions may result in infeasible plans (e.g., collisions between robots), highlighting its interdependency.
    \item \textbf{Slice and Serve} is an extended task of "Slice Food" by adding another goal of serving the vegetable/fruit slices in a desired location (e.g., a bowl or plate).
\end{tightlist}
We created $10$ problems for Pick and Place, Slice Food, and Slice and Serve. For the Pick Obstacle tasks, we created $9$ problems common to Single Arm and Dual Arm. All problems are provided with PDDL problems, linguistic instructions, and scene observations. We also created a PDDL domain shared across the tasks.

\begin{figure}
\vspace{-2mm}
\begin{minipage}{\linewidth}
\begin{algorithm}[H]
  \caption{ViLaIn-TAMP Planning and Execution}
  \label{alg:algorithm}
  \small
  \begin{algorithmic}[1]
    \Require $\mathcal{D}$, $\mathcal{S}$, $\mathcal{L}$, $N_{\text{CP,max}}$
    \State $\mathcal{O}, \mathcal{I}, \mathcal{G} \gets \textproc{ViLaIn-INITIAL}(\mathcal{D}, \mathcal{S}, \mathcal{L})$
    \State $P_{\text{initial}} \gets \{\mathcal{O}, \mathcal{I}, \mathcal{G}\}$
    \State $\pi, \mathcal{E} \gets \textproc{TAMP}(P_{\text{initial}})$
    \State $\mathcal{H}_{\mathcal{P}} \gets [\,]$, $\mathcal{H}_{\mathcal{E}} \gets [\,]$
    \State $N_{\text{CP}} \gets 0$
    \While{$\pi \neq \textproc{SUCCESS}$ \textbf{and} $N_{\text{CP}} < N_{\text{CP,max}}$}
        \State $N_{\text{CP}} \gets N_{\text{CP}} + 1$
        \State $P_{\text{revised}} \gets \textproc{ViLaIn-REVISE}(P_{\text{initial}}, \mathcal{E}, \mathcal{H}_{\mathcal{P}}, \mathcal{H}_{\mathcal{E}})$
        \State $\mathcal{H}_{\mathcal{P}}.\textit{append}(P_{\text{revised}})$, $\mathcal{H}_{\mathcal{E}}.\textit{append}(\mathcal{E})$
        \State $\pi, \mathcal{E} \gets \textproc{TAMP}(P_{\text{revised}})$
    \EndWhile
    \If{$\pi = \textproc{SUCCESS}$}
        \State \textproc{Execute-Real-Robot}($\pi$)
    \EndIf
  \end{algorithmic}
\end{algorithm}
\end{minipage}
\figuretextspace
\end{figure}

\subsubsection{\textbf{Baselines and Comparisons}}
To evaluate our framework and show the effectiveness of corrective planning (CP), we compare the following methods:
\begin{tightlist}
    \item \textbf{\vilaintamp-CP} is our proposed framework \uline{with} CP. 
    The maximum number of CP attempts is controlled by $N_{\text{CP, max}}$ (e.g., $N_{\text{CP, max}}=3$ allows CP for three times).
    \item \textbf{\vilaintamp-No-CP} is our proposed framework \uline{without} CP.
    \item \textbf{Baseline-CP} is a VLM-as-a-planner baseline~\cite{mei2024repanvlm, yang2025guiding} that uses LLMs/VLMs to directly generate action plans \uline{with} CP, taking $\mathcal{O}$, $\mathcal{L}$, $\mathcal{D}$ as input and directly generating {$\mathcal{A}$} as output, which are then solved by the TAMP module (denoted hereon as the \textit{baseline approach}).
\end{tightlist}
We note that end-to-end approaches such as RT-X~\cite{oneill2024openx} and VIMA~\cite{zhu2023vima} are not directly comparable in our setting, as they do not support PDDL-based symbolic reasoning or pre-execution geometric verification. 
Since \vilain instantiates a similar formalizer paradigm as LLM+P~\cite{liu2023llm}, the \vilaintamp-No-CP variant is a LLM+P-style baseline, isolating the effect of the corrective feedback loop; AutoTAMP~\cite{chen2024autotamp} instead targets navigation and is not directly applicable.

\begin{figure*}[t]
    \centering
    \includegraphics[width=0.875\linewidth]{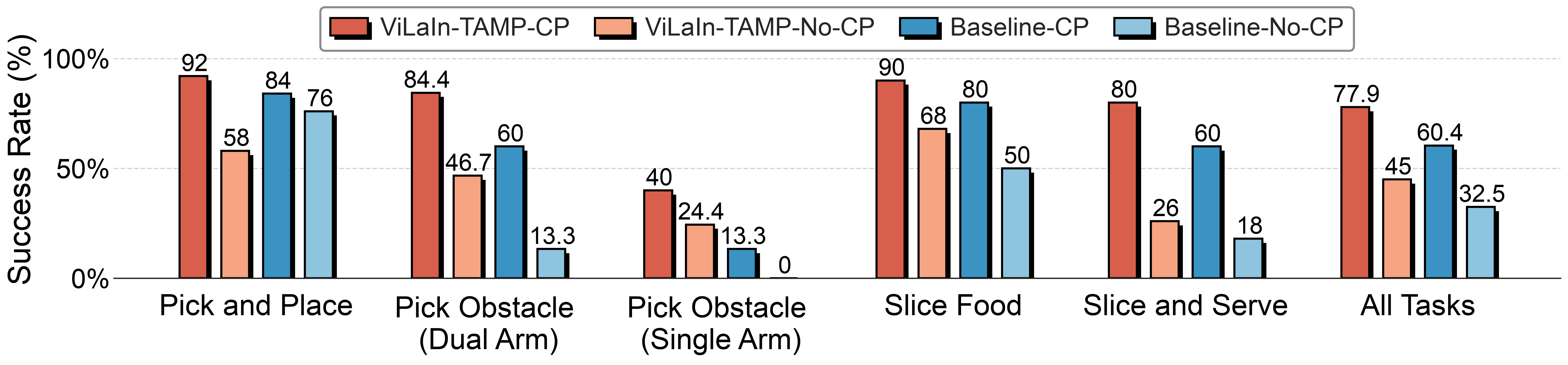}
    \vspace{-2mm}
    \caption{\textbf{Comparison of \vilaintamp and the baseline,} evaluating their performance on five cooking tasks with and without corrective planning (CP). 
    The maximum number of CP attempts is set to $3$. \vilaintamp consistently outperforms the baseline in all tasks. CP is effective in both models, consistently improving the success rate by a large margin.
    } 
    \label{fig:main-results}
    \figuretextspace 
\end{figure*}

\subsubsection{\textbf{Evaluation Metrics}}
We employ the success rate (\%) as our main evaluation metric. A trial is considered successful if (1) a symbolically complete, geometrically feasible (i.e., collision-free) plan is found, and (2) the plan is executed successfully, achieving the target goal. 

\subsubsection{\textbf{Foundation Model Choices}}
\vilaintamp consists of foundation models. In our experiments, we adopted Qwen-2.5VL-7B-Instruct~\cite{bai2025qwen2} for the object estimator\footnote{Due to poor object detection by GPT-4o in preliminary experiments, we used Qwen-2.5VL.} and GPT-4o~\cite{openai2024gpt4o} for the other modules and the corrective planning. 
In object detection, we assumed that the positions of fixed objects (e.g., robot arms and the cutting board) appearing throughout the tasks are known and only detected objects (foods and other tools) that change depending on the task. 

\subsubsection{\textbf{Planning Setup}}
\cref{alg:algorithm} shows \vilaintamp's planning and execution process. We first generate an initial problem $P_{\text{initial}}$ using \vilain, which is passed to the TAMP module. If planning fails, \vilain revises $P_{\text{initial}}$ based on failure feedback $\mathcal{E}$, maintaining a history of revisions $\mathcal{H}_{\mathcal{P}}$ and feedbacks $\mathcal{H}_{\mathcal{E}}$ for context. Planning succeeds if a complete plan $\pi$ is found within $N_{\text{CP, max}}$ attempts.

\subsection{Primary Results}
\label{subsec:primary_results}

\textbf{Comparison against Baseline ($\mathcal{Q}$1).} As shown in~\cref{fig:main-results}, \vilaintamp-CP achieves the highest success rate across all tasks. Compared to the baseline, \vilaintamp outperforms the VLM-as-a-planner approach by an average margin of 17.5\%, with even more significant improvements observed as task complexity increases. 
In~\cref{table1}, we report planning-related metrics for each of the tasks, including average task planning attempts, average motion planning attempts, and planning time statistics (mean and standard deviation), denoted as SR, TP, MP, Mean, and SD, respectively.
\vilaintamp is shown to have stable planning durations with relatively low variance. We attribute the high deviations in certain tasks, such as Slice and Serve, to the increase in task complexity and longer horizons (e.g., slicing up to two objects). 

\begin{figure}
    \centering
    \includegraphics[width=1.0\linewidth]{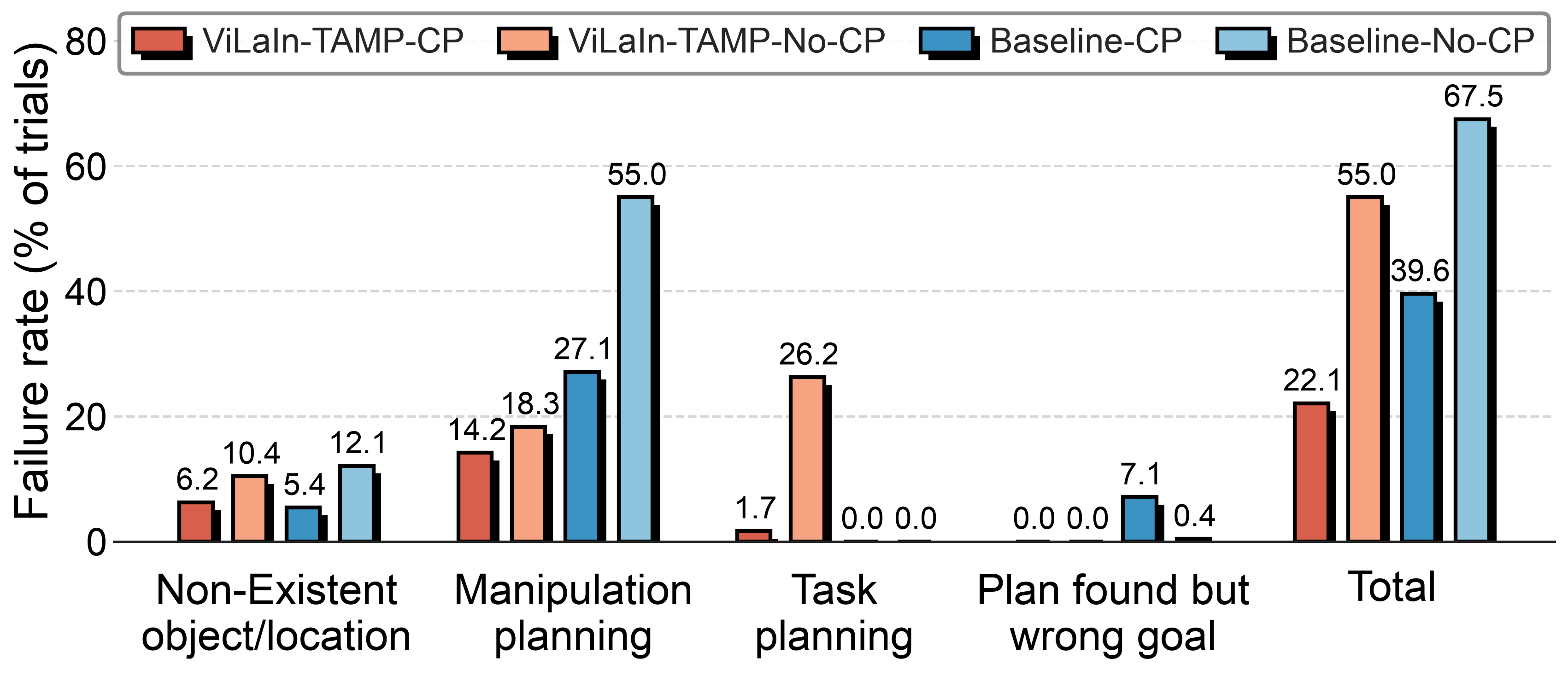}
    \caption{Failure modes as a rate over all trials.}
    \label{fig:failure_modes}
    \figuretextspace
\end{figure}

\textbf{Failure Modes Analysis ($\mathcal{Q}$2).} 
As shown in~\cref{fig:failure_modes}, failure modes have been classified as the following: (1) detecting non-existing objects/locations, (2) manipulation planning failure, (3) task planning failures, and (4) successful plans but wrong goal. In mode (1), the model often misrecognizes objects or locations (e.g., an apple is detected as a tomato). A more accurate object detection model may reduce these failures. In mode (2), manipulation planning failures occur due to collision with other objects or between robots. Results show \vilaintamp reduces this failure type versus the baseline. In some cases, however, the model cannot resolve motion failures, as it misinterprets whether a condition or its negation should be applied. In mode (3), task planning often fails due to incomplete, incorrect, or contradictory use of the predicates. In mode (4), the model may generate a valid task and motion plan for unintended tasks (e.g., the instruction is to move an object to the cutting board, but it moves the object to the plate). Failure mode (4) occurs significantly in the baseline method. In contrast, \vilaintamp, which leverages a symbolic task planner, did not experience such hallucinations, demonstrating the importance of symbolic representation for increasing reliability and predictability.

\begin{figure}
    \centering
    \scalebox{1.0}{
    \includegraphics[width=0.9\linewidth]{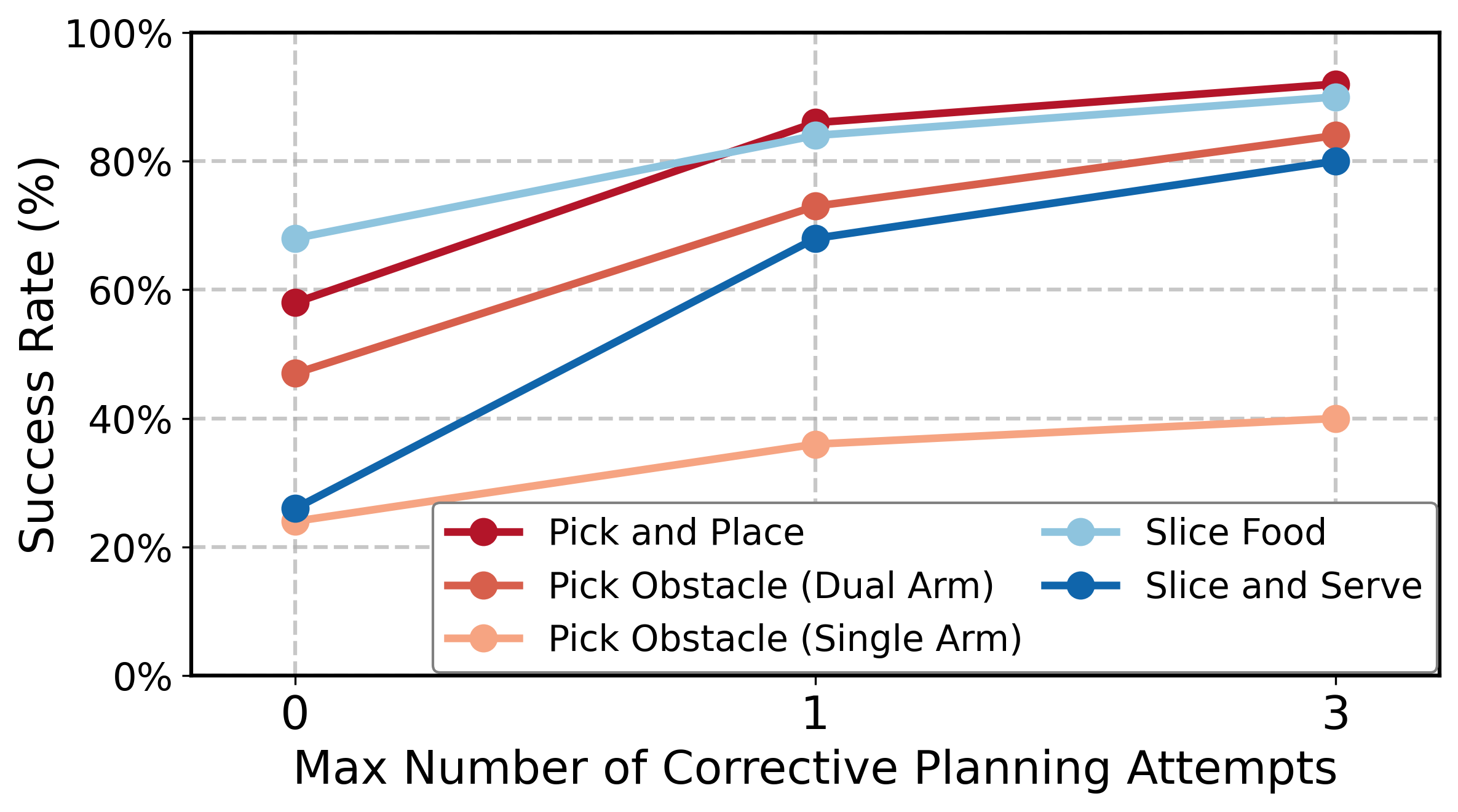}
    }
    \caption{\textbf{Effectiveness of Corrective Planning for \vilaintamp-CP.} 
    More CP attempts consistently improve the success rate.
    }
    \label{fig:cp_results}
    \figuretextspace  

\end{figure}

\subsection{Ablation Studies}
\begin{table*}[!t]
    \centering
    \caption{Additional Metrics for \vilaintamp with and without in-context learning (ICL)}
    \footnotesize
        \begin{tabular}{lccccccccccc}%
            \toprule
            \multicolumn{1}{c}{\multirow{3}{*}{\textbf{Task Name}}} &
            \multicolumn{5}{c}{\textbf{Without ICL}} &
            \multicolumn{5}{c}{\textbf{With ICL = 1}} \\
            \cmidrule(lr){2-6}\cmidrule(lr){7-11} &
            \multicolumn{3}{c}{\textbf{Metrics}} &
            \multicolumn{2}{c}{\textbf{Planning Times [sec]}} &
            \multicolumn{3}{c}{\textbf{Metrics}} &
            \multicolumn{2}{c}{\textbf{Planning Times [sec]}} \\
            \cmidrule(lr){2-4}\cmidrule(lr){5-6}
            \cmidrule(lr){7-9}\cmidrule(lr){10-11} & 
            \textbf{\%SR} & \textbf{\#TP} & \textbf{\#MP}& 
            \textbf{Mean} & \textbf{SD}&
            \textbf{\%SR} & \textbf{\#TP} & \textbf{\#MP}&
            \textbf{Mean} & \textbf{SD}\\
            
            \midrule
            Pick and Place& 92.0 & 1.64 & 1.24 & \phantom{0}6.24 & \phantom{0}0.11 & 92.0 (+0.0) & 1.50 & 1.30 & \phantom{0}6.27 & \phantom{0}0.14 \\
            Pick Obstacle Dual & 84.4 & 2.00 & 1.64 & 11.11 & \phantom{0}2.72 & 89.0 (\textcolor{ForestGreen}{+4.6}) & 1.56 & 1.44 & \phantom{0}9.59 & \phantom{0}1.66 \\
            Pick Obstacle Single & 40.0 & 2.76 & 1.60 & 13.89 & \phantom{0}3.32 & 49.0 (\textcolor{ForestGreen}{+9.0}) & 2.60 & 1.62 & 11.75 & \phantom{0}4.95  \\
            Slice Food & 90.0 & 1.60 & 1.28 & 16.41 & \phantom{0}1.70 & 98.0 (\textcolor{ForestGreen}{+8.0}) & 1.46 & 1.30 & 16.40 & \phantom{0}1.61  \\
            Slice and Serve & 80.0 & 2.30 & 1.80 & 32.37 & 10.97 & 86.0 (\textcolor{ForestGreen}{+6.0}) & 2.02 & 1.67 & 30.30 & 12.31 \\
            
            \bottomrule
            
        \end{tabular}
    \label{table1}
\end{table*}

\begin{figure*}[t]
    \centering
    \includegraphics[width=0.95\linewidth]{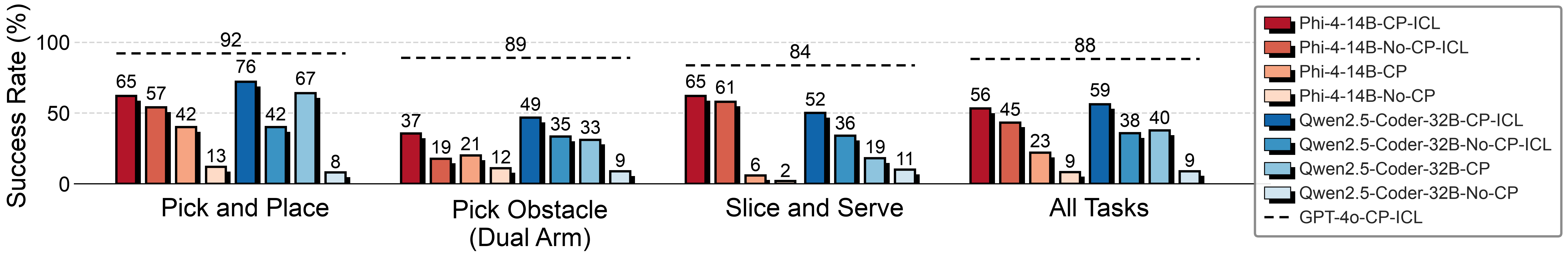}
    \vspace{-4mm}
    \caption{\textbf{Evaluation of Open-Source Models.} The results of GPT-4o-CP-ICL are plotted for reference.}
    \label{fig:open_source_results}
    \figuretextspace 
\end{figure*}

\textbf{Effect of Corrective Planning ($\mathcal{Q}$3).} We verify the effect of corrective planning by using different maximum corrective planning attempts ($N_{\text{CP, max}} = 0, 1, 3$). Overall, the results show that ViLaIn-TAMP-No-CP has a lower average success rate of 45\%. CP significantly improves the success rate of all tasks (\cref{fig:cp_results}).

\textbf{Effect of In-Context Learning ($\mathcal{Q}$4).} We analyze whether providing ICL input-output examples has a significant impact on \vilaintamp's performance in \cref{table1}. 
Overall, we observe a 5.5\% increase in the mean success rate for \vilaintamp-CP. Using ICL examples proves to be more significant in the case of \vilaintamp-No-CP, as the mean success rate increases from 45\% to 55\%.
The results also show that adding the ICL examples reduces the average task planning attempts across all tasks, which suggests that the examples help in reducing incorrect or contradictory propositions, increasing consistency, and the success rate of initial generations.

\textbf{Choice of Foundation Models ($\mathcal{Q}$5).} In recent years, there has been significant progress in the development of open-source foundation models. Motivated by this, we evaluate performance when those models are used instead of GPT-4o. We select two open-source models: Microsoft's Phi-4~\cite{microsoft2024phi4} (14B parameters) and Alibaba's Qwen2.5-Coder-32B-Instruct~\cite{hui2024qwen25codertechnicalreport} (32B parameters). Note that these models are text-only and that Qwen2.5-VL-7B-Instruct is used as the object estimator. 
We select $3$ tasks for evaluation: Pick and Place, Pick Obstacle Dual Arm, and Slice and Serve. We generate $5$ outputs for each problem and evaluate the resulting $50$ trials per task ($45$ for Pick Obstacle Dual Arm). \cref{fig:open_source_results} shows results. The results of GPT-4o-CP-ICL, which is our strongest model from~\cref{subsec:primary_results}, are shown for reference. We can see that both CP and ICL consistently improve success rates across all tasks, with CP+ICL exceeding $50$\% on every task, confirming their effectiveness in open-source models as well. On the other hand, there is still about $30$\% performance gap compared to GPT-4o, suggesting that it remains difficult to achieve competitive performance with open-source models.

\subsection{Real Robot Validation}
\label{subsec:real_robot_validation}

\begin{figure}
    \centering
    \includegraphics[width=0.825\linewidth, trim={0 1cm 0 2cm}, clip]{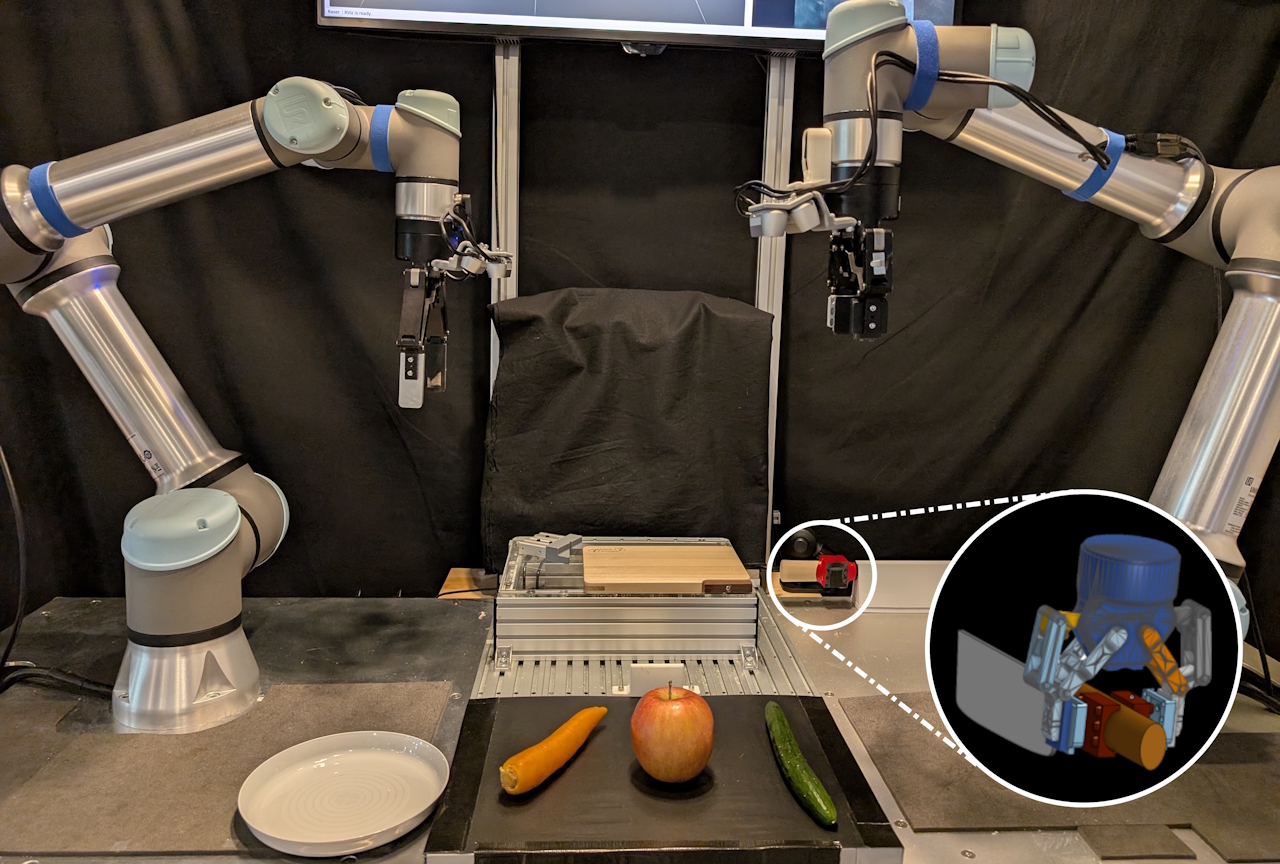}
    \caption{\textbf{Overview of our robotic system.}
    Dual arm robotic system for real-world validation on a cooking domain. }
    \label{fig:physical_setup}
    \figuretextspace  
\end{figure}

\textbf{Real Robot Setup:} Our physical robot system comprises two robotic arms (UR5e). Each arm has an incorporated force-torque sensor at its wrist and is equipped with a wrist-mounted RGB-D camera (RealSense D435) and a parallel gripper (2F-85 and 2F-140). The arm allocated for the slicing action has a Robotiq 2F-85 parallel gripper with custom fingertips to equip the kitchen knife, as shown in \cref{fig:physical_setup}.

\textbf{Real-World Perception:} 
During planning, TAMP uses a rough estimation of the object poses, but real-world execution requires a perception module to estimate the object poses accurately. We combine the Qwen2.5-VL-7B-Instruct open-vocabulary detector~\cite{bai2025qwen2} with the Segment Anything Model 2~\cite{ravi2024sam} for object detection and segmentation. 

\textbf{Pre-defined Skill Library:} 
In \vilaintamp, we build a library of skills for subtasks involving complex, contact-rich interactions. For instance, in the Slice Food task, slicing actions rely on Reinforcement Learning (RL) with compliance control. Prior work \cite{beltran2024sliceit} showed this approach adapts to unseen objects and can capture an object's force profile with a single slice. As formalized in \cref{subsec:tamp}, each skill is a black-box operator with symbolic pre/post conditions and object-centric start/end poses; the RL policy is short-horizon, and execution halts if the ending pose is not reached. The library can be further extended by training additional skills in the same way.

\textbf{Real-World Execution:} We begin with a valid plan generated by \vilaintamp. This plan is considered \textit{optimistic} because, although it is based on current visual observations, it may not fully capture positional uncertainties of objects or account for environmental changes caused by the robot's actions (for example, the altered state of an object after slicing). To ensure robust and reliable execution in the real world, we adopt the following approach:

\begin{tightlist}
    \item \textbf{Action Classification:} Post-planning, actions are classified as:
    \textbf{(A)} Trackable motion-based actions (e.g., pick, place, equip tool).
    \textbf{(B)} Environment-altering learning skills (e.g., slicing).
    
    \item \textbf{Plan Segmentation:} The plan is segmented according to this classification, grouping only \textbf{(A)} actions for potential replanning.
    
    \item \textbf{Adaptive Execution:} During execution, actions in \textbf{(B)} are carried out sequentially and independently to update our understanding of the environment. Before executing any group of \textbf{(A)} actions, we replan them if needed based on the latest state, ensuring the plan remains robust to any changes or uncertainties.
\end{tightlist}

\textbf{Real-world validation} confirmed the overall feasibility of ViLaIn-TAMP on a physical dual-arm system, though two main failure modes were observed. The most frequent involved object displacement during slicing, leaving objects in poses that the subsequent grasping actions could not recover from. A secondary challenge was grasping deformable objects: vegetable slices can be easily squeezed and become slippery, making open-loop grasping unreliable.

\section{Conclusion}
\label{sec:conclusion}
This work introduced \vilaintamp, a hybrid planning framework that integrates VLMs with symbolic and geometric planning to enable robust, interpretable, and autonomous bimanual robot manipulation in real-world environments. Using \vilain to translate multimodal input into PDDL problems and a TAMP system to find action plans, our framework systematically verifies the logical and physical feasibility of the generated plans. Across five challenging bimanual cooking tasks, \vilaintamp outperformed VLM-as-a-planner baselines with increased interpretability, especially as task complexity increases. The corrective planning module, which iteratively refines plans based on grounded failure feedback from our custom MTC implementation, proved essential for robustness. In future work, \vilaintamp can be extended with automatic PDDL domain generation~\cite{oswald2024large} and visual failure reasoning to further reduce manual effort and improve error recovery.

%%%%%%%%%%%%%%%%%%%%%%%%%%%%%%%%%%%%%%%%%%%%%%%%%%%%%%%%%%%%%%%%%%%%%%%%%%%%%%%%


%%%%%%%%%%%%%%%%%%%%%%%%%%%%%%%%%%%%%%%%%%%%%%%%%%%%%%%%%%%%%%%%%%%%%%%%%%%%%%%%

%%%%%%%%%%%%%%%%%%%%%%%%%%%%%%%%%%%%%%%%%%%%%%%%%%%%%%%%%%%%%%%%%%%%%%%%%%%%%%%%

% \section*{ACKNOWLEDGMENT}

%%%%%%%%%%%%%%%%%%%%%%%%%%%%%%%%%%%%%%%%%%%%%%%%%%%%%%%%%%%%%%%%%%%%%%%%%%%%%%%%

{
    \bibliographystyle{IEEEtran}
    \bibliography{root}
}

\end{document}